\pdfoutput=1

\documentclass[11pt]{article}

\usepackage[preprint]{acl}

\usepackage{times}
\usepackage{latexsym}

\usepackage[T1]{fontenc}

\usepackage[utf8]{inputenc}

\usepackage{microtype}

\usepackage{inconsolata}

\usepackage{graphicx}
\usepackage{subfigure}
\usepackage{amsmath,amsfonts}
\usepackage{algorithmic}
\usepackage{algorithm}
\usepackage{multirow}
\usepackage{caption}
\usepackage{subcaption}
\usepackage{color}

\title{Skip-Thinking:  Chunk-wise Chain-of-Thought Distillation Enable Smaller Language Models to Reason Better and Faster}

\author{Xiaoshu Chen,  Sihang Zhou,  Ke Liang, Xiaoyu Sun, Xinwang Liu \\
        National University of Defense Technology}

\begin{document}
\maketitle
\begin{abstract}
Chain-of-thought (CoT) distillation allows a large language model (LLM) to guide a small language model (SLM) in reasoning tasks. Existing methods train the SLM to learn the long rationale in one iteration, resulting in two issues: 1) Long rationales lead to a large token-level batch size during training, making gradients of core reasoning tokens  (i.e., the token will directly affect the correctness of subsequent reasoning) over-smoothed as they contribute a tiny fraction of the rationale. As a result, the SLM converges to sharp minima where it fails to grasp the reasoning logic. 2) The response is slow, as the SLM must generate a long rationale before reaching the answer. Therefore, we propose chunk-wise training (CWT), which uses a heuristic search to divide the rationale into internal semantically coherent chunks and focuses SLM on learning from only one chunk per iteration. In this way, CWT naturally isolates non-reasoning chunks that do not involve the core reasoning token (e.g., summary and transitional chunks) from the SLM learning for reasoning chunks, making the fraction of the core reasoning token increase in the corresponding iteration.
Based on CWT, skip-thinking training (STT) is proposed. STT makes the SLM automatically skip non-reasoning medium chunks to reach the answer, improving reasoning speed while maintaining accuracy.
We validate our approach on a variety of SLMs and multiple reasoning tasks.
\end{abstract}

\section{Introduction}
\begin{figure*}[t]
\centering
  \setlength{\belowcaptionskip}{-0.5cm}
  \includegraphics[width=0.9\linewidth]{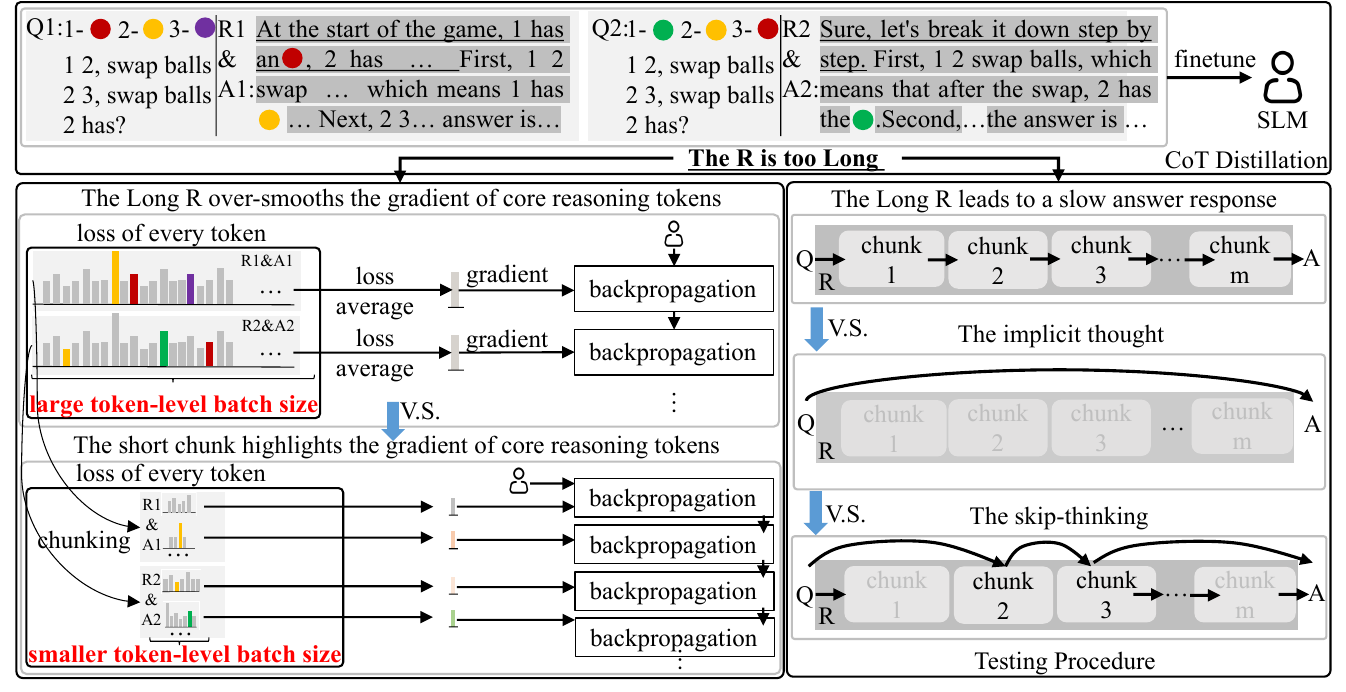}
  \caption{Illustration of CoT Distillation. The batch size is set to 1 as an illustrative example. The core reasoning token (like the yellow and green ball in rationale R) means that its accuracy can determine the subsequent reasoning process. 1) \textit{Superficial understanding}: The large token-level batch size will cause the gradient of the core reasoning token to be over-smoothed by plenty of other non-reasoning tokens (highlighted with a gray background in R) that are similar across different rationales during backpropagation, leading to SLMs converging to a sharp minimum where SLM often makes mistakes when generating the core reasoning token. 2) \textit{Time-consuming}: Generating the full R takes longer than outputting the answer A directly.}
  \label{fig:motivation}
\end{figure*}

Chain of Thought (CoT) \citep{chu2023survey} distillation enables small language models (SLMs) \citep{gpt2,t5} to replicate the reasoning patterns of large language models (LLMs) \citep{instructgpt,llama2, llama3}, enhancing their reasoning abilities for domain-specific tasks.
The training procedure for mainstream CoT distillation methods \citep{ho-etal-2023-large,magister2022teaching,renzhu2022specializing} is shown in the top box of Figure \ref{fig:motivation}. It requires the SLM to learn a long reasoning process (rationale) from the LLM for a given task in a single training iteration, leading to two problems.

\textbf{1) Superficial understanding}. The training loss for the SLM is computed as the average value over all target tokens. Consequently, the token-level batch size corresponds to the number of training tokens within a mini-batch. Since the rationale is long, the token-level batch size remains large even with a batch size of 1. Large batch size typically causes gradient over-smoothing during backpropagation \citep{bs2018finding,bslarge,bs}, thereby leading to a generalization gap. Specifically, the model updates with the average gradient of the tokens in the batch. As the batch size increases gradually—consider an extreme case where a single batch encompasses the entire training dataset—the gradients across batches become more similar, causing the model loss to decrease rapidly along the similar gradients and converge to a sharp minimum.
More critically, in CoT distillation, the core reasoning tokens (such as the yellow and green ball in the rationale of Figure \ref{fig:motivation}) constitute a small proportion of rationales, while the prevalence of similar non-reasoning tokens (e.g., those used for transition and summarization) across different rationales exacerbates gradient over-smoothing, causing the SLM to converge rapidly towards learning the expressive patterns of the LLM rather than core reasoning logic.


\textbf{2) Time-consuming.} The SLM trained with these methods requires completing the full rationale to produce the final answer during testing, resulting in a significantly slower response time.

To address the first problem, some naive approaches, such as weighting the loss of core reasoning tokens or prompt LLMs to remove redundant expressions in rationale, do not perform well (see Appendix \ref{sec:naive}). In this work, we propose a chunk-wise training (CWT) strategy. CWT utilizes a chunking data generator that introduces a heuristic search guided by model loss to segment the rationale into a fixed number of internal semantically coherent chunks and focus SLMs on learning from one chunk per iteration. 
By doing so, the token-level batch size is smaller, which mitigates gradient over-smoothing. More importantly, since certain non-reasoning chunks function solely as preludes, summaries, or transitions (like the underlined text in rationale in Figure \ref{fig:motivation}), and thus lack core reasoning tokens, when the SLM learns reasoning chunks independently within a given iteration, CWT naturally isolates the influence of non-reasoning chunks on core reasoning token learning (see Appendix \ref{sec:nonreasoningstas}), thereby directing the SLM's attention toward learning the core reasoning logic during that iteration.

For the second question, several methods \citep{hsieh-etal-2023-distilling, MMI, deng2023implicitchainthoughtreasoning, deng2024explicit} have been proposed to enhance the response speed of the answer. Among them, internalizing the explicit reasoning process \cite{deng2023implicitchainthoughtreasoning, deng2024explicit} into the latent space has emerged as a promising direction. However, these internalization-based methods may compromise answer accuracy due to the lack of an explicit reasoning process. Similar to these approaches, we hypothesize that language models can also encode explicit reasoning within the latent space. 

However, we argue that, akin to humans who externalize parts of their reasoning to maintain coherence and mitigate forgetting key information, language models should externalize the reasoning chunks that contain core reasoning tokens to facilitate subsequent reasoning. Therefore, we propose a CWT-based skip-thinking training (STT) strategy. Specifically, STT uses answer correctness as a criterion to determine whether internalizing a specific reasoning chunk is reasonable. If the answer remains correct after removing the chunk, this chunk is deemed non-essential and can be internalized from the output. Otherwise, the chunk should be externalized during reasoning. In this way, STT constructs training data that makes the SLM automatically skip unimportant non-reasoning chunks to accelerate the response while still arriving at the correct answer. 

The key contributions are as follows:

1) To prevent a superficial understanding, we provide a theoretical analysis from the perspective of gradient updates and propose the CWT to enhance SLMs' capability in comprehension of reasoning logic.

2) The STT is proposed based on reasoning internalization, which not only preserves reasoning accuracy but also accelerates SLM reasoning.

3) Plenty of experiments are conducted across 3 different SLMs and 7 reasoning tasks to verify our proposed method.

\section{Related works}
CoT \citep{chu2023survey} is first introduced by \citet{wei2022chain}. Subsequently, CoT distillation and reasoning acceleration emerges as two critical research directions aimed at broadening the application scope of CoT.

\subsection{CoT distillation}
CoT distillation is first introduced in concurrent works by \citet{ho-etal-2023-large}, \citet{magister2022teaching}, and \citet{renzhu2022specializing}. They prompt the LLM to generate rationales for a given task, which is then applied as the supervised label to make the SLM mimic the reasoning logic of the LLM. Building upon these works, Scott \citep{wang-etal-2023-scott} is introduced to enhance the alignment of the SLM's rationale with the answer. \citet{li2023turning} proposes integrating the LoRA \citep{hu2022lora} to enhance the utilization of negative samples generated by the LLM. PaD \citep{zhu2023pad} employs an external code compiler to enhance the performance of the SLM. In addition to the aforementioned work on improving the distillation mechanism, some works have integrated CoT distillation with information retrieval \citep{PTR}, table reasoning \citep{table}, thereby broadening the application scope of CoT distillation.

However, the aforementioned methods enable the SLM to learn the full rationale for the given task in a single iteration, which may cause the SLM to superficially understand the reasoning logic of LLMs.

\subsection{CoT acceleration}
The existing methods to accelerate the reasoning process can be roughly divided into three directions: multi-task learning, post-thinking mechanism, and latent space thought. 

Multi-task learning \citep{hsieh-etal-2023-distilling, MMI, liu-etal-2024-minds} utilizes distinct prefixes to differentiate between tasks. For instance, when the input task prefix is \textit{[label]}, the SLM directly outputs the answer, whereas when the input task prefix is \textit{[rationale]}, the SLM outputs the rationale. Since multi-task learning allows for outputting the answer directly, the answer response time can align with that of the standard fine-tuning that only applies the answer to train SLM. However, because the rationale and the answer are not within the same output sequence, the conclusion of the SLM's rationale often fails to align with the answer directly output by the SLM.

Post-thinking mechanism \citep{chen2024distillingreasoningabilitylarge} trains the SLM to output the rationale after providing the answer, so that the answer can be generated first during the test. However, the post-thinking sacrifices the ability to decompose the task through the rationale, making it more challenging to handle tasks with higher complexity. 

\begin{figure*}[ht]
\centering
    \setlength{\belowcaptionskip}{-0.5cm}
  \includegraphics[width=0.8\linewidth]{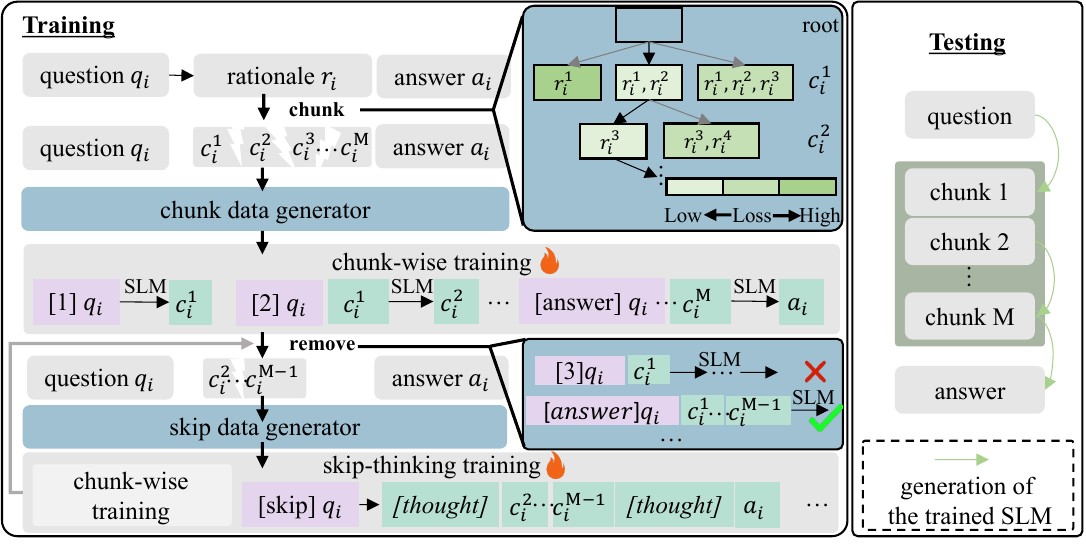}
  \caption{The illustration of the proposed methods. The flames indicate that the model is undergoing training, and the \textit{[thought]} is a specail token that represents the SLM is thinking in mind.}
  \label{fig:method}
\end{figure*}

Training SLMs to reason in latent space has emerged as a recent research direction \citep{deng2023implicitchainthoughtreasoning, goyal2024pasue, deng2024explicit, hao2024latent}. These methods propose internalizing explicit rationales into latent space, enabling implicit reasoning during forward propagation to directly generate answers. For instance, \citet{deng2024explicit} gradually removes reasoning steps during training to internalize rationales, while \citet{hao2024latent} introduces a special token, \textit{[thought]}, to facilitate latent reasoning. However, this approach may reduce answer accuracy in some tasks compared to explicit reasoning. We posit that this stems from the model’s tendency to forget previous reasoning steps during extended reasoning in the latent space. Explicit rationales serve as a scratchpad that facilitates problem-solving \citep{wei2022chain}. When discarded, the model is more likely to forget prior steps, leading to degraded reasoning capacity.

\section{Preliminary}
Let $D=\{(q_i, a_i) | i=0,1,...,n\}$ refer to the original dataset consisting of $n$ samples for training SLM, where $q_i$ and $a_i$ represent the question and answer, respectively. Based on $D$, CoT distillation first utilizes a zero-shot or few-shot CoT prompt to make LLM output rationale $r_i$ for $q_i$. Then, the SLM is trained to maximize the generation
likelihood of $r_i$ and $a_i$. The training loss per training iteration of CoT distillation can be formulated as:
\begin{equation}
    \mathbb{L}= \frac{1}{B}\sum_{b=0}^{B}\frac{1}{K-s}\sum_{k=s}^{K} \ell(f_{\vartheta}(x^{b}_{1,k}, ), x^{b}_{k+1})
\label{eq1}
\end{equation}
where $x = q \oplus r \oplus a$ is the input sequence whose length is $K$ ($\oplus$ refers to the string concatenation), $B$ refers to the training batch size, $s$ is the start index of $r \oplus a$ in $x$, $f_{\vartheta}(\cdot)$ represents the forward calculation of SLM with parameters $\vartheta$, and $\ell(\cdot)$ is the cross-entropy loss. After training, SLM has the ability to think before outputting answers.

\textbf{Superficial understanding}. Considering the parameters of SLM as a whole, during backpropagation, the gradient of $\vartheta$ can be expressed as:
\begin{equation}
    \frac{\partial \mathbb{L}}{\partial \vartheta} = \frac{\partial}{\partial \vartheta} \left( \frac{1}{N} \sum_{i=1}^N \ell_i \right) = \frac{1}{N} \sum_{i=1}^N \frac{\partial \ell_i}{\partial \vartheta}
\label{eq2}
\end{equation}
where $N=B\times (K-s)$ represents the token-level batch size and $\ell_i$ is the cross-entropy loss for $i_{th}$ training token. Assume that we divide the training tokens into two sets $S_1$ and $S_2$, where $S_1$ is the training token set involving the core logic in any single reasoning step and $S_2$ is the remaining tokens, Equation \ref{eq2} can be rewritten as:
\begin{equation}
    \frac{\partial \mathbb{L}}{\partial \vartheta} = \frac{1}{N} \sum_{i=1}^{|S_1|} \frac{\partial \ell_i}{\partial \vartheta} + \frac{1}{N} \sum_{j=1}^{|S_2|} \frac{\partial \ell_j}{\partial \vartheta}
\label{eq3}
\end{equation}
Since $|S_2|$ is usually much larger than $|S_1|$, the gradient of the token in $S_1$ will be smoothed by the gradient of the token in $S_2$, which ultimately leads to a superficial understanding of SLM in this reasoning step.

\textbf{Slow answer response}. The SLM trained according to Equation \ref{eq1} must first generate a rationale before providing an answer, which leads to a slower answer response compared to the SLM that directly outputs the answer.

\section{Method}
To address the two problems, we propose CWT and STT. Figure \ref{fig:method} illustrates the process. First, the LLM generates $r_i$ for $q_i$. Details on obtaining $r_i$ are in Appendix \ref{sec:generation}. The chunk and skip data generators then sequentially generate data for CWT and STT. 
\subsection{Chunk data generator}
The chunk data generator divides the complete rationale into smaller chunks and makes SLMs learn from each chunk independently during a single training iteration. After the division, $|S2|$ in Equation \ref{eq3} is significantly reduced in the iteration of learning reasoning chunks, allowing SLMs to concentrate on comprehending the essential reasoning logic within the given chunks.

Division methods vary in granularity: sentence-level, reasoning step-level, and chunk-level. The first two methods lead to duplicate generation due to task-specific variations in sentence and reasoning step numbers (see Appendix \ref{sec:sentandstep}). Chunk-level division segments the rationale into $M$ chunks. Training SLM with this data will make the SLM reach the answer after $M$ distinct stages, thereby avoiding duplicate generation. Thus, the chunk-level division is employed in the chunk data generator.

\subsubsection{Average chunking}
When performing chunking, the simplest way is to divide the reasoning steps into $M$ parts equally. Specifically, we first split the rationale by \textit{"$\backslash n$"} to obtain $r_i=\{r_i^{j} | j=0,1,..L\}$ that has $L$ reasoning steps. Then the reasoning steps contained in the $m_{th}$ chunk can be formulated as:
\begin{equation}
    c_i^m = \begin{cases}
\{r_i^{j} \mid j \in [g\times m, g\times m + g)) \} & m < M
\\
\{r_i^{j} \mid j \in [g\times m, L] \}& m=M
\end{cases}
\label{eq4}
\end{equation}
where $g = \lfloor L/M \rfloor$ and $ j \in \mathbb{Z} $. After chunking, we can convert a training sample $x_i$ into $M+1$ training data. The first $M$ training data can be formalized as:
\begin{equation}
\{[m] \oplus q_i \oplus c_i^{1} \oplus c_i^{2} ... \oplus c_i^{m} | m = 0, 1, ..., M\}
\label{eq5}
\end{equation}
and the $M+1$ training sample is:
\begin{equation}
[answer] \oplus q_i \oplus c_i^{1} \oplus c_i^{2} ... \oplus c_i^{M} \oplus a_i
\label{eq6}
\end{equation}
The reason for adding the prefix $[m]$ and $[answer]$ is that it can tell the model what stage the current reasoning is at, thereby reducing the difficulty of reasoning. And the $s$ in the Equation \ref{eq1} is the start index of $c_i^{m}$ and $a_i$ in these data at this time.

\begin{algorithm}[t]
    \setlength{\belowcaptionskip}{-0.9cm}
    \caption{Search-based chunking}
    \begin{algorithmic}[1] 
        \REQUIRE Chunk list $c_{ij}$ of $r_i$ and SLM $\vartheta_{j}$ before $(j+1)$ training epoch, $q_i$, $a_i$, threshold $\eta$, $M$
        \FOR{$m$ in range($M - 1$)}
            \STATE Calculate the loss $l_c$ for $c_{ij}^{m}$ with Equation \ref{eq1}
            \STATE Merge $c_{ij}^{m}$ and $c_{ij}^{m+1}$ to form the list $c_{temp}$
            \STATE Initial: $l_{min}$ $\leftarrow$ $+\infty$, $index$ $\leftarrow$ $+\infty$
            \FOR{$idx$ in range(len($c_{temp}$))}
            
                \STATE Calculate the loss $l_{idx}$ for $c_{temp}[:idx]$ with Equation \ref{eq1}
                \IF {$l_{idx} < l_{min}$}
                    \STATE $l_{min}$ $\leftarrow$ $l_{idx}$, $index$ $\leftarrow$ $idx$
                \ENDIF
            \ENDFOR
            \IF {$l_c - l_{min}  > \eta$ }
                \STATE $c_{ij}^{m}$ $\leftarrow$ $c_{temp}[:index]$
                \STATE $c_{ij}^{m+1}$ $\leftarrow$ $c_{temp}[index:]$
            \ENDIF
        \ENDFOR
        \ENSURE  Chunk list $c_{i(j+1)}$ of $r_i$
    \end{algorithmic}
\label{algorithm}
\end{algorithm}

\subsubsection{Search-based chunking}
Since the average chunking (AC) may divide multiple semantically coherent reasoning steps into different chunks, the reasoning fluency of the SLM may degrade after training. In addition, the combinatorial space for allocating $L$ reasoning steps to $M$ chunks is vast. Therefore, we propose a search-based chunking (SBC) that applies the loss of SLM as heuristic information to efficiently identify a better chunking result.

The detailed process of SBC is outlined in Algorithm \ref{algorithm}. The initial chunking result $c_i^0$ is obtained through AC. Algorithm \ref{algorithm} is executed before each training epoch. In general, the loss of the language model on the target token sequence indicates the language model's understanding of the content within the target token sequence \citep{wan2024knowledge}. Based on this point, in Algorithm \ref{algorithm}, we progressively increase the number of reasoning steps allocated to the current searching chunk and compute the SLM loss for it. As the loss decreases, we infer that the reasoning steps allocated to this chunk are more comprehensible to the SLM, aiding its understanding of the information in the current reasoning stage. Thus, we utilize this loss comparison as heuristic information to iteratively adjust the chunk division with a greedy strategy, reducing suboptimal results from unreasonable division during training.

\subsection{Skip data generator}
To accelerate reasoning, we employ a skip data generator for STT. STT is essentially to internalize the rationale. However, unlike \citet{deng2024explicit}, STT still requires the explicit output of the SLM to provide a clear intermediate basis for subsequent reasoning.


Specifically, the skip data generator sequentially removes chunk and uses the SLM trained with CWT to predict the answer. Taking $c_{i}^{m}$ as a example of the removed chunk, the answer prediction process is initialized with the input $[m+1] \oplus q_i \oplus c_{i}^{0} \oplus \cdots \oplus c_{i}^{m-1}$ and proceeds until the model produces an answer. If the answer is incorrect, this indicates that the removed chunk contains key reasoning information that should be externalized during the reasoning process. Otherwise, it suggests that the current chunk is non-essential—likely serving as a transitional or summary component—and its contribution can be internalized by the model. After this chunk-removal procedure, the skip data generator produces training data illustrated in Figure \ref{fig:method} for STT to build an SLM capable of skipping the non-reasoning chunk.
It is important to note that 1) before STT, the SLM is initialized using the original pre-trained parameters, rather than those fine-tuned by CWT, reducing the risk of overfitting; and 2) CWT remains incorporated into STT training, ensuring that the SLM is still exposed to the full rationale during training. Furthermore, as shown by the grey arrow in Figure \ref{fig:method}, the above process can be iterated until the reasoning accuracy of the SLM no longer increases.

\subsection{Testing}
After training the SLM with the CWT and STT, when prompting SLM with the input $[skip] \oplus q_{test}$, the SLM can adaptively skip the unimportant reasoning chunk and only externalize the key reasoning chunks, thereby accelerating reasoning while ensuring reasoning accuracy.

\section{Experiments}
We first introduce the detailed experimental settings, followed by a series of experiments to validate the following aspects. \textbf{Q1:} The effect of each proposed module on the model's answer accuracy. \textbf{Q2:} Comparison between the proposed method and the state-of-the-art method. \textbf{Q3:} Can CWT indeed mitigate the superficial understanding issue in SLM? \textbf{Q4:} The distinction between skip-thinking and full-thinking.

\subsection{Experimental setting}

Seven reasoning benchmarks, categorized into four distinct types: arithmetic, symbolic, common sense, and other logical reasoning, are employed to evaluate our method. Detailed information about the datasets can be found in Appendix \ref{sec:dataset}. For conciseness, we denote each dataset using abbreviations derived from their concatenated initials.

Unless otherwise stated, LLM in this section refers to \textit{text-davinci-002} 175B, developed based on InstructGPT \citep{instructgpt} and accessible via the OpenAI API. As for the student SLM, we employ GPT-2 (ranging from the base to large model) \cite{gpt2} and T5 (ranging from the small to large model) \cite{t5} to evaluate the effectiveness of the proposed methods. The detailed generation parameters for LLMs and SLMs are given in Appendix \ref{sec:generation}. More training details are available in the Appendix \ref{sec:traindetail}.

\begin{table*}[ht]
\renewcommand\arraystretch{1.2}
\resizebox{\linewidth}{16mm}{
\begin{tabular}{ccccccccccccccc}
\hline
\multirow{2}{*}{Methods} & \multicolumn{7}{c}{GPT2-base}          & \multicolumn{7}{c}{T5-small}           \\ \cline{2-15} 
                         & SE & AD & MA & Svamp & TSO & LLC & SQA & SE & AD & MA  & Svamp & TSO & LLC & SQA \\ \hline
Base                     &8.55&10.08&14.44&10.66& 56.88&21.33&58.22&3.94&8.40& 8.88 &  9.00& 60.00&  45.33 &56.04     \\
Base w. AC               &7.89&10.92&16.11& 8.66& 97.32&24.66&58.80&3.94&7.56& 8.88 & 10.00& 65.33&  46.66 &57.20\\
Base w. SBC              &8.55&10.92&17.77&11.33&100.00&25.33&59.38&4.60&8.40& 9.33 & 10.33& 99.55&  48.66 &57.78\\
Base w. SkipALL          &7.89&11.76&17.22&11.00&100.00&11.33&59.97&3.94&8.40& 8.88 &  9.66& 99.55&  28.66 &58.80\\
Base w. STT             &\textbf{10.52}&\textbf{12.60}&\textbf{22.77}&\textbf{12.33}&\textbf{100.00}&\textbf{28.00}&\textbf{60.55}&\textbf{5.92}&\textbf{10.08}&\textbf{11.66} &\textbf{11.33}&\textbf{99.55}& \textbf{48.66}  &\textbf{59.97}\\ \hline
\end{tabular}
}
\caption{The accuracy of various methods across different datasets. Refer to the Appendix \ref{sec:extablation} for additional ablation experiments using various student SLMs.}
\label{tab:1}
\end{table*}

\begin{table*}[ht]
\setlength{\belowcaptionskip}{-0.3cm}
\renewcommand\arraystretch{1.2}
\resizebox{\linewidth}{22.5mm}{
\begin{tabular}{ccccccccccccccc}
\hline
Methods & SE                   & AD                   & MA                &  Svamp                  & TSO                  & LLC                  & SQA                  & SE                   & AD                   & MA                & Svamp                   & TSO                  & LLC                  & SQA                  \\ \hline
Text-davinci-002 (175B)         &81.50	&76.71	&78.79		&	64.20	&	53.20	&57.71		&	53.45  &81.5	&76.71	&78.79		&	64.20	&	53.20	&57.71		&	53.45                    \\
 \hline
 & \multicolumn{7}{c}{GPT2-base (124M)}          & \multicolumn{7}{c}{T5-small(60M)}           \\ \hline 
Standard finetune        &8.55&10.08&14.44&10.66& 56.88&21.33&58.22&3.94&8.40& 8.88 & 9.00& 60.00&  45.33 &56.04                        \\ \hline
CoT-Finetuing            & 8.55 &	10.08&	14.44	&		10.66	&	56.88&	21.33	&		58.22& 3.94&8.40& 8.88 & 9.00&60.00&  45.33 &56.04 \\
Scott *                    & 9.21	&9.24&	22.22&			11.33&		56.44&	22.00	& 55.74 & 5.26 & 7.56 & 10.00 & 10.33 & 70.22 & 46.00 & 58.36 \\ \hline
Step-by-Step             & 7.89	&\textbf{12.60}&	17.22&			10.00	&	94.66&	4.00	&		59.67  &  2.63&	8.40&	10.55	&		8.33	&	99.11	&25.33	&		58.36                 \\
MMI                     &- &-&-&-&-&-&-&3.28&	7.56&	10.00	&		10.33	&	\textbf{99.55}	&25.33	&		57.78 \\
ICoT-SI                & 2.63 & 4.20 & 4.33 & 3.88 & 36.00 &0.00  &52.40  & - & - & - & - & - & - & - \\ \hline
Ours                &\textbf{10.52}&\textbf{12.60}&\textbf{22.77}&\textbf{12.33}&\textbf{100.00}&\textbf{28.00}&\textbf{60.55}&\textbf{5.92}&\textbf{10.08}&\textbf{11.66} &\textbf{11.33}&\textbf{99.55}& \textbf{48.66}  &\textbf{59.97}\\ \hline
\end{tabular}
}
\caption{A comparison of our methods with other approaches. A dash (-) indicates that the official code of the method is not implemented on the corresponding SLM. An asterisk (*) indicates that Scott requires the complete logits of each output token for implementation; thus, the rationales used in Scott are collected from the open-source model LLama3.1-70b-instruction \citep{llama3}.}
\label{tab:2}
\end{table*}

\subsection{Ablation experiments for Q1}
 \textbf{First,} we conduct a series of comprehensive experiments to assess the effectiveness of each proposed strategy. The results are presented in Table  \ref{tab:1}. Further experiments involving SLMs with varied parameters as student models are detailed in the Appendix \ref{sec:extablation}. The baseline model adopts the full-thinking training approach proposed by \citet{ho-etal-2023-large}. 

It is evident that when chunks are partitioned using the AC, the performance of the SLM improves relative to the baseline across most tasks. But in a few tasks, the model's performance declines. We attribute this to the AC dividing coherent reasoning steps into separate chunks, thereby reducing SLM reasoning coherence. Thus, when a more optimal SBC is applied for chunking, the SLM exhibites improved performance across all tasks.

\begin{figure}[t]
\centering
    \setlength{\abovecaptionskip}{-0.1cm}
    \setlength{\belowcaptionskip}{-0.5cm}
  \includegraphics[width=\linewidth]{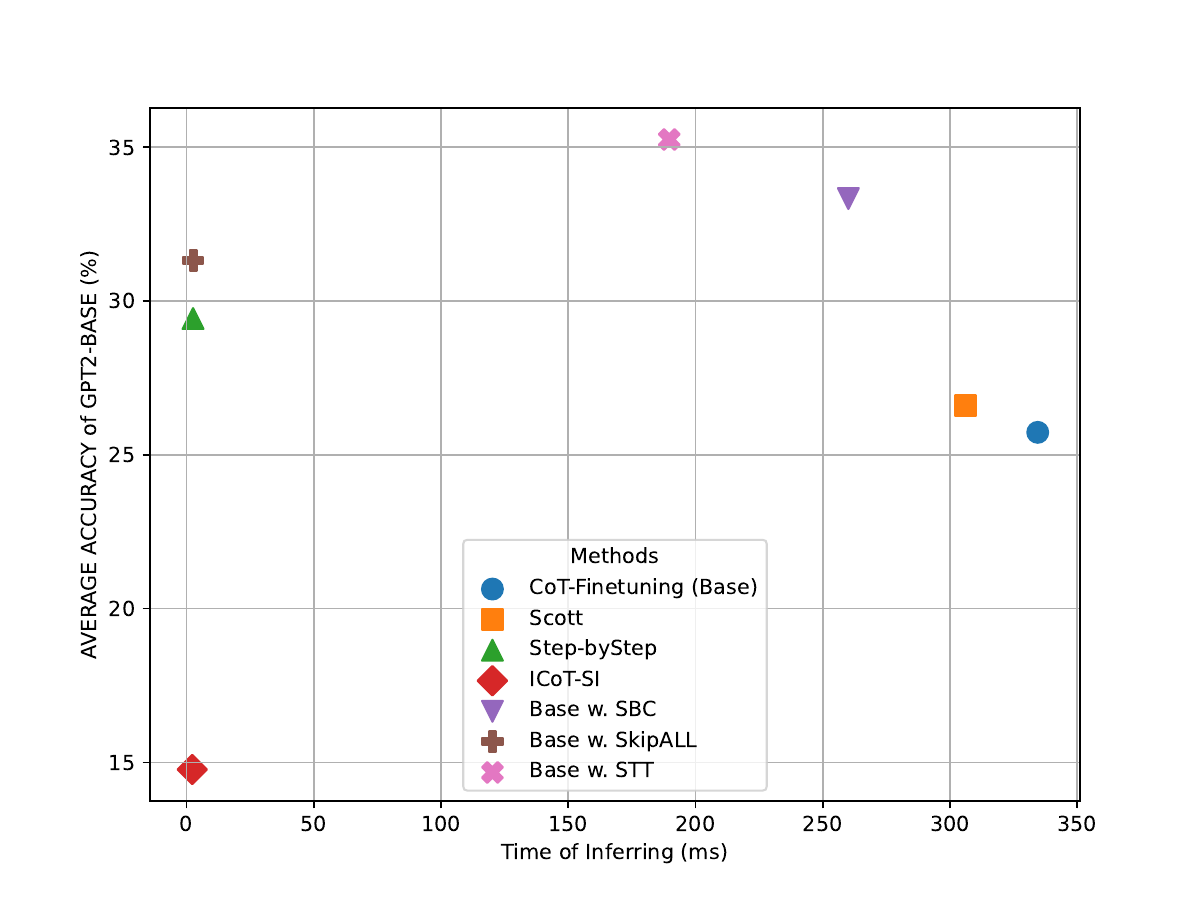}
  \caption{A comprehensive comparison of the average inference speed and performance across different methods on all datasets using GPT2-base.}
  \label{fig:speed}
\end{figure}

Building upon the SBC, we additionally apply STT to train the SLM. To further clarify the effectiveness of STT, we implement a variant referred to as \textit{Base w. SkipALL}. This variant does not consider the answer as the judgment criterion during training data construction for STT, but instead trains the SLM to directly bypass all intermediate reasoning steps. Experimental results show that this variant leads to a notable decline in SLM performance, especially for the LLC dataset. We attribute the significant performance decline of the variant on the LLC dataset to the fact that the LLC dataset requires parallel reasoning rather than sequential reasoning, where each reasoning step is independent with no context dependence between them. Therefore, when using the variant, the SLM needs to reason about multiple different subtasks in parallel in the latent space, which is hard for SLMs and leads to the decline in performance.

In contrast to this variant, \textit{Base w. STT} achieves a consistent performance improvement, highlighting the benefit of externalizing parts of the reasoning process to preserve key information. We also observe that, compared to \textit{Base w. SBC}, \textit{Base w. STT}, which restricts output to key reasoning chunks, also shows improved performance. We attribute this to only retaining essential reasoning chunks lowers the risk of SLM hallucinations—a point we discuss in more detail in section \ref{skip-thinking}.

\textbf{Then, we verify the impact of different chunk numbers $M$ on SBC.} In Figure \ref{fig:cnum}, we can observe that for tasks with relatively fixed reasoning methods and steps, such as common sense and symbolic reasoning, the SLM works best when $M$ is close to the average number of reasoning steps $L$. For mathematical reasoning, which has a large variation in reasoning methods and steps, setting $M$ greater than $L$ helps the SLM learn more solutions, thereby improving the performance of SLMs.

\textbf{Third, the comparison of chunking result between AC and SBC} are shown in Appendix \ref{sec:SBCcase}, which intuitively proves that SBC can better make the reasoning steps within a chunk more coherent.


\subsection{Comparison with Other Methods for Q2}
\begin{table}[t]
\setlength{\belowcaptionskip}{-0.65cm}
\resizebox{\linewidth}{11mm}{
\begin{tabular}{cccc}
\hline
Method                                           & Token type            & AD(\%)                   & TSO(\%)                  \\ \hline
\multirow{2}{*}{Base}                            & core reasoning tokens & 87.37                &   89.25                   \\
                                                 & other tokens          & 89.64                &   95.18                   \\ \hline
\multicolumn{1}{l}{\multirow{2}{*}{Base w. SBC}} & core reasoning tokens & 88.88  & 92.73 \\
\multicolumn{1}{l}{}                             & other tokens          & 89.80  & 95.17 \\ \hline
\end{tabular}
}
\label{tab:3}
\caption{Confident score of GPT2-base for different tokens.}
\end{table}

The comparison methods include standard finetuning (using only answers as label), few-shot prompting for LLMs (specific prompts can be found in the \citet{ho-etal-2023-large}), full-thinking CoT distillation (CoT-Finetuning \citep{ho-etal-2023-large}, Scott\cite{wang-etal-2023-scott}), and distillation methods that accelerate SLM inference via multi-task learning(step by step \citep{hsieh-etal-2023-distilling}, MMI \citep{MMI}) and internalized chains of thought (ICoT-SI \citep{deng2024explicit}. 

As shown in Table \ref{tab:2}, our proposed method outperforms the other distillation approaches and achieves performance close to that of LLMs on certain tasks. Although the inference speed remains slower than that of multi-task learning and internalized chains of thought, it strikes a balance between performance and inference speed (see Figure \ref{fig:speed}). Finally, we present a comparison of our method against other baselines in terms of training time and GPU memory consumption in Appendix \ref{compare_other}, demonstrating that our method requires less GPU memory and does not spend too much additional training time.

\subsection{Validate CWT for Q3}
\begin{figure}[ht]
\centering
\setlength{\abovecaptionskip}{-0.1cm}
    \setlength{\belowcaptionskip}{-0.5cm}
  \includegraphics[width=\linewidth]{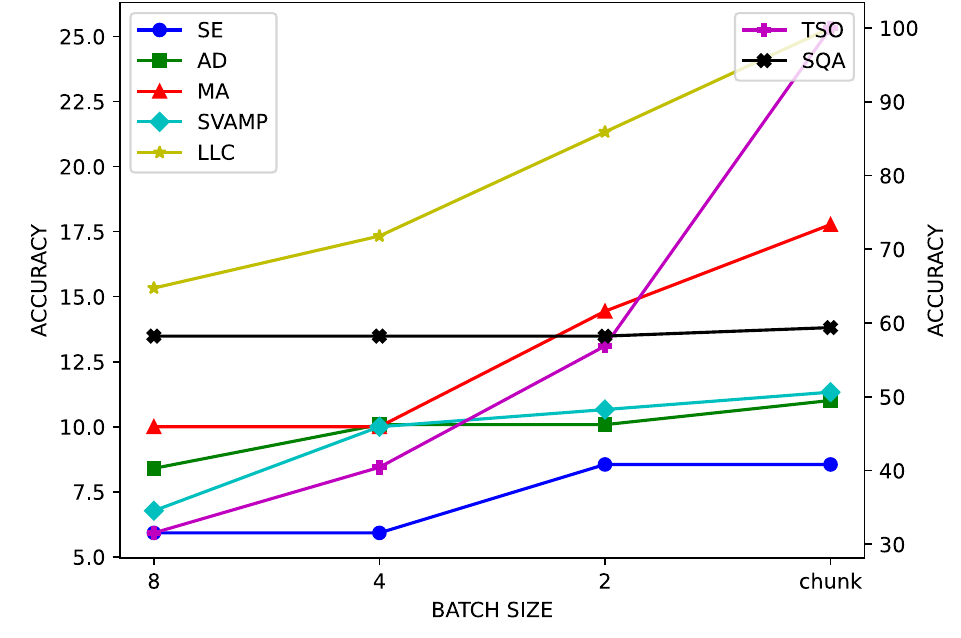}
  \caption{GPT2-base's performance trend when the batch size changes. Batch size is proportional to token-level batch size. Chunk means using CWT with SBC.}
  \label{fig:batchsize}
\end{figure}

\textbf{First, we show the performance of SLM as the token-level batch size changes in Figure} \ref{fig:batchsize}. It can be seen that as the token-level batch size decreases, the performance of SLM on various reasoning tasks increases, which strongly verifies the motivation of CWT, that is, a smaller token-level batch size helps SLM converge to a flat minimum.

\begin{figure}[t]
\centering
    \setlength{\belowcaptionskip}{-0.2cm}
  \includegraphics[width=\linewidth]{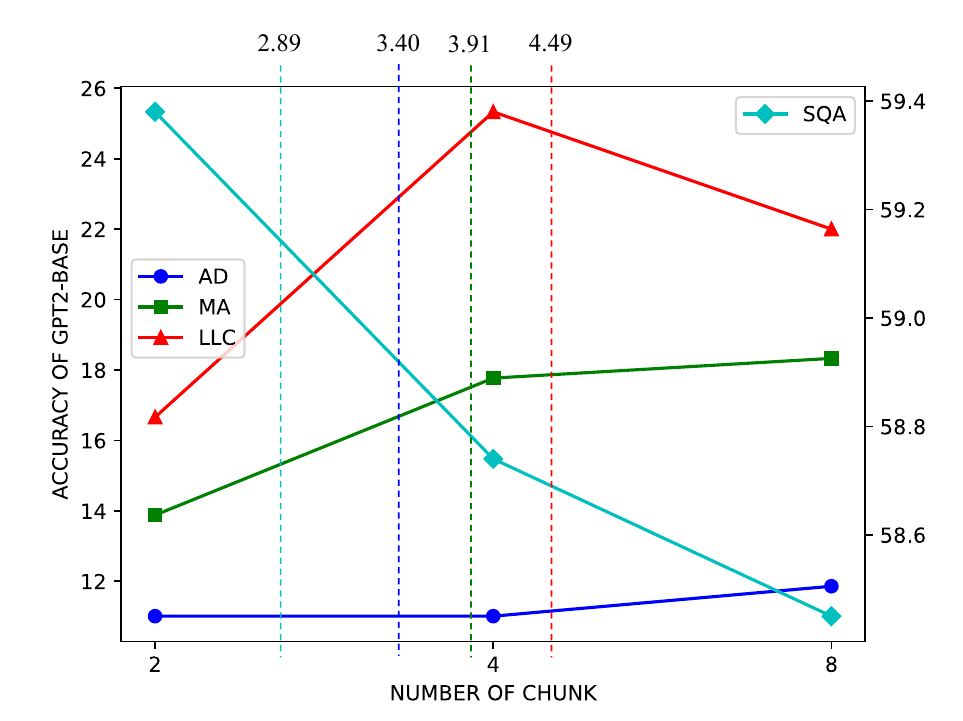}
  \caption{SLM performance trend when the number of chunks changes. The vertical dotted line refers to the average number of reasoning steps.}
  \label{fig:cnum}
\end{figure}

\textbf{Subsequently, we further verify whether CWT helps SLM learn the core reasoning logic.} Specifically, mathematical expressions (in AD) and key exchange results (in TSO) are identified and extracted as core reasoning tokens. Then, we counted the average confidence score of the core reasoning tokens and the non-reasoning tokens when the trained SLM output rationale. One can observe that compared with the base model, the gap between the confidence score of the core reasoning tokens and that of the common tokens is smaller after using CWT, which means that the SLM with CWT is more confident when outputting the core reasoning tokens, i.e., it better understands the core reasoning logic of the current task.

\textbf{Then, we show the cases} (Appendix \ref{sec:corecase}) where the correct answer is inferred after using CWT compared to base because the core reasoning token is predicted correctly. This also proves that CWT helps SLMs comprehend the core reasoning logic.

\textbf{Finally, the reasoning speed of the SLM trained with CWT based on SBC is faster than that of the baseline}, which can be observed in Figure \ref{fig:speed}. We argue that this improvement stems from the SLM trained with the former focusing more on the correctness of the reasoning logic and exhibiting greater conciseness in its reasoning expressions. This conciseness is reflected in the length of the generated rationale. The average number of words in the rationale generated by the former across all tasks is 50, while the latter generates 56 words.

\subsection{Validate Skip-thinking for Q4} 
\label{skip-thinking}
In addition to verifying the speed-accuracy trade-off of skip-thinking shown in Figure \ref{fig:speed}, we conduct two additional experiments.

\textbf{Reasoning acceleration} Skip-thinking can automatically skip unimportant chunks, leading to faster inference compared to full-thinking. We also present the acceleration ratio of skip-thinking relative to full-thinking across different datasets in the Table \ref{tab:sscrate}. We observe the following: 1) skip-thinking yields inference speedup across datasets; and 2) the degree of acceleration varies across dataset types. For more complex math problems, since more key information needs to be output, skip-thinking skips fewer chunks, resulting in less acceleration compared to simpler tasks such as commonsense question answering (SQA) or object-swap reasoning (TSO). In the case of LLC, since it requires decomposition into multiple subtasks with no inter-task dependency and each task only involves one step reasoning, skip-thinking retains almost the entire reasoning process for each subtask, resulting in inference latency comparable to full-thinking.

\textbf{Case study.} The Appendix \ref{sec:core_skip} presents some case studies, demonstrating the advantage of skip-thinking over full-thinking. By omitting intermediate reasoning steps, skip-thinking is less susceptible to model output hallucinations.


\begin{table}[t]
\setlength{\belowcaptionskip}{-0.5cm}
\centering
\resizebox{1\linewidth}{5.3mm}{
\renewcommand\arraystretch{1.2}
\renewcommand\arraystretch{1.2}
\begin{tabular}{cccccccc}
\hline
 & SE & AD & SVAMP & MA & TSO & LLC & SQA  \\ \hline
SBC / STT & 1.29 & 1.32 & 1.38 & 1.33 & 1.89 & 1.08 & 1.57   \\ \hline
\end{tabular}
}
\caption{Reasoning speedup ratio of STT compared to SBC on GPT2-base.}
\label{tab:sscrate}
\end{table}

\section{Conclusion}
When using full rationale for CoT distillation, SLM faces two challenges: superficial understanding and slow response times. To address the two problems, we first propose CWT to reduce the token-level batch size, enhancing SLM's reasoning by mitigating gradient over-smoothing. To maintain coherence, a chunking method based on heuristic search to divide rationale into semantically coherent blocks is introduced. Building on CWT, STT trains SLM to adaptively skip the non-reasoning chunks. Leveraging CWT and STT, the SLM achieves faster and more accurate reasoning.

\section*{Limitations}
\label{limitation}
SBA employs a greedy search strategy, which may result in identifying only locally optimal chunk modes rather than globally optimal ones. For this point, strategies such as simulated annealing can be employed to avoid local optima (see Appendix \ref{sec:localoptima}).

\section*{Ethics Statement}
Given that toxicity is present in LLMs, the student SLM may inherit such toxicity during the learning of the LLM's reasoning process. To address this issue, one can apply existing toxicity reduction techniques to mitigate toxicity in LLM reasoning.


\bibliography{custom}

\appendix

\section{Experimental Details}
\label{sec:expdetail}

\subsection{Datsets}
To evaluate our model, we employ seven established benchmarks spanning four categories: Arithmetic (SingleEq \citep{singleeq}, AddSub \citep{addsub}, MultiArith \citep{mutilarith}, Svamp \citep{svamp}), Symbolic (Last Letter Concatenation \citep{NEURIPS2022_8bb0d291}), Common Sense (StrategyQA \citep{strategyqa}), and General Logical Reasoning (Track Shuffled Objects \citep{data_}). We implement the training-test data partitioning adhering to the methodology described by \citep{ho-etal-2023-large}.
\label{sec:dataset}

\subsection{Rationale generation of \textit{Text-davinci-002}}
\label{sec:generation}
We utilize the prompts described in \citet{ho-etal-2023-large} to generate rationales from \textit{Text-davinci-002}. The key modification involves swapping the positions of the rationale and the answer in the few-shot exemplars, enabling the LLM to leverage the answer information during reasoning. In alignment with the methodology outlined by \citet{ho-etal-2023-large}, we constrain the teacher-generated rationales to a maximum sequence length of 128. Additionally, we employ temperature sampling with T=0.7 to generate diverse rationales for each sample.

\subsection{Rationale generation of SLM}
The student model predictions are limited to a sequence length of 1024 and greedy decoding is applied for SLM across all benchmarks.

\subsection{Training datails}
For SLM training, we configure a batch size of 2, an initial learning rate of 1e-5, and a total of 50 epochs. We evaluate the SLM after each epoch. The learning rate follows a cosine annealing schedule with restarts, incorporating a warm-up phase of 1200 steps. We employ the Adam optimizer with hyperparameters  \(\beta_1 = 0.9\), \(\beta_2 = 0.95\), and \(weight\_decay = 0.1\) to optimize the model parameters. For search-based chunking, we set $\eta=0.1$, as this value can empirically promote stable model training. As for the number of chunks $M$, We assign $M=4$ for all arithmetic reasoning tasks and Last Letter Concatenation, and $M=2$ for Track Shuffled Objects and StrategyQA. The effect of different $M$ on SLM performance is shown in Figure \ref{fig:cnum}.

\label{sec:traindetail}

\begin{table}[t]
\renewcommand\arraystretch{1.2}
\begin{tabular}{p{\linewidth}}
\hline
[instruction] Please output strictly according to the format of Example.
[example] Question: Alice, Bob, and Claire are playing a game. At the start of the game, they are each holding a ball: Alice has a orange ball, Bob has a purple ball, and Claire has a pink ball. As the game progresses, pairs of players trade balls. First, Alice and Claire swap balls. Then, Bob and Alice swap balls. Finally, Alice and Claire swap balls. At the end of the game, Alice has the Which choice is true? Answer choices: (A) purple ball. (B) orange ball. (C) pink ball.Why the answer is B.
Explanation: 
1. Alice-orange, Bob-purple, Claire-pink ball.
2. Alice-pink, Bob-purple, Claire-orange.
3. Alice-purple, Bob-pink, Claire-orange.
4. Alice-orange, Bob-pink, and Claire-purple.
Question:$\{ \#$ question$\}$ Why the answer is $\{ \#$ Answer$\}$
Explanation: \\ \hline
\end{tabular}
\caption{The prompt for concise rationale.}
\label{tab:naiveprompt}
\end{table}

\begin{table*}[ht]
\renewcommand\arraystretch{1.2}
\begin{tabular}{p{\linewidth}}
\hline
Question: Alyssa picked 17 plums and Jason picked 10 plums . Melanie picked 35 pears . How many plums were picked in all ? \\

Rationale: Alyssa picked 17 plums. Jason picked 10 plums. 17 + 10 = 27 plums. Melanie picked 35 pears. 27 + 35 = 62 There were 62 fruits picked in all. \\ \hline

chunk 1: Alyssa picked 17 plums. Jason picked 10 plums. 17 + 10 = 27 plums. \\ \hline

chunk 2: Melanie picked 35 pears. 27 + 35 = 62. \\ \hline

chunk 3: There were 62 fruits picked in all. \\ \hline
\end{tabular}
\caption{Analysis of chunk result. Since chunk 3 is just a summary statement, the average proportion of core reasoning tokens in reasoning chunks (chunk 1 and 2) is greater than that in the complete rationale.}
\label{tab:chunk result Analysis}
\end{table*}

\begin{table}[t]
\centering
\renewcommand\arraystretch{1.2}
\begin{tabular}{ccc}
\hline  
                    &  Full Rationale  &   Reasoning Chunks        \\  \hline
Proportion          &  8.93 \text{\%}  &   12.16 \text{\%}       \\  \hline
\end{tabular}
\caption{Comparison between the proportion of core reasoning tokens in the reasoning chunk and that in the complete rationale.}
\label{tab:rate comparasion}
\end{table}

\begin{table}[t]
\centering
\resizebox{\linewidth}{15mm}{
\renewcommand\arraystretch{1.2}
\begin{tabular}{cccc}
\hline  
\multicolumn{4}{c}{GPT2-base (124M)} \\ \hline
                         & Base & Base w. Weight & Base w. Refine   \\ \hline
TSO                     &  37.33  &   36.88 & 43.11       \\  \hline
\multicolumn{4}{c}{GPT2-medium (355M)}    \\ \hline
TSO                     &  41.77  &   42.22 & 36.88       \\  \hline
\end{tabular}
}
\caption{The accuracy of different methods on TSO. Base refers to \citet{ho-etal-2023-large} without diverse rationale. Base w. Weight and Base w. Refine represent the two naive solutions to address the oversmoothing problem.}
\label{tab:naive}
\end{table}

\begin{table*}[t]
\renewcommand\arraystretch{1.2}
\resizebox{\linewidth}{29.5mm}{
\begin{tabular}{ccccccccccccccc}
\hline
\multirow{2}{*}{Methods} & \multicolumn{7}{c}{GPT2-medium (355M)}          & \multicolumn{7}{c}{T5-base (220M)}           \\ \cline{2-15} 
                         & SE & AD & MA & SVAMP & TSO & LLC & SQA & SE & AD & MA & Svamp & TSO & LLC & SQA \\ \hline
Base                     &11.84&15.96&18.88&10.00& 67.11&24.00&59.24& 6.57&10.92&17.22& 10.66&71.11&64.66&54.87    \\
Base w. AC               &11.18&17.64&17.22&10.00& 76.00&26.66&60.64& 8.55&11.76&16.11&12.66& 77.33&74.00&60.11     \\
Base w. SBC              &12.50&19.32&19.44&10.66& 87.11&28.66&61.13& 9.21&13.44&17.77&\textbf{13.33}& 93.33&79.33&60.98     \\
Base w. SkipALL          &11.84&19.32&18.88&10.33&\textbf{100.00}&14.66&61.42& 8.55&12.60&17.77&11.66& 99.55&39.33&62.01     \\
Based w. STT             &\textbf{15.78}&\textbf{21.01}&\textbf{20.00}&\textbf{11.33}&\textbf{100.00}&\textbf{30.66}&\textbf{62.15}& \textbf{10.52}&\textbf{15.96}&\textbf{19.44}&\textbf{13.33}&\textbf{100.00}&\textbf{82.66}&\textbf{62.44}     \\ \hline
\multicolumn{1}{c}{Methods} & \multicolumn{7}{c}{GPT2-large (774M)}          & \multicolumn{7}{c}{T5-large(700M)}           \\ \hline
Base                     &13.15&15.96&20.00&11.00& 68.88&25.33&60.84&  9.21&14.28&17.77&12.33& 92.44&76.66&57.64   \\
Base w. AC               &12.50&16.80&21.11&12.33& 85.77&27.33&61.57&  8.55&15.96&13.88&13.33& 95.08&81.33&61.71 \\
Base w. SBC              &16.44&17.64&23.33&14.00& 94.66&28.66&62.44& 10.52&16.80&19.44&14.00&100.00&82.66&63.75     \\
Base w. SkipAll          &15.78&17.64&21.66&13.33&100.00&14.66&62.88&9.86&15.96&18.88&13.66&100.00&54.66&63.75     \\
Based w.STT              &\textbf{17.10}&\textbf{19.32}&\textbf{24.44}&\textbf{15.00}&\textbf{100.00}&\textbf{30.66}&\textbf{63.31}&\textbf{12.50}&\textbf{17.64}&\textbf{21.11}&\textbf{14.66}&\textbf{100.00}&\textbf{85.33}&\textbf{64.04}     \\ \hline
\end{tabular}
}
\caption{The performance of SLM under different models and different training strategies}
\label{tab:extablationbase}
\end{table*}

\begin{table}[t]
\centering
\renewcommand\arraystretch{1.6}
\resizebox{0.92\linewidth}{11mm}{
\begin{tabular}{cccccc}
\hline  
\multicolumn{6}{c}{Llama3.2-1B} \\ \hline
                         & Base & Base w. AC & Base w. SBC & Base w. SkipALL & Base w. STT  \\ \hline
GSM8K                     &  52.23  & 49.88 & 54.66 & 52.38 &   \textbf{55.26}     \\  \hline
\multicolumn{6}{c}{Llama3.2-3B}    \\ \hline
GSM8K                     &  79.15  &   77.25 & 79.52 & 79.30 &  \textbf{80.21}     \\  \hline
\end{tabular}
}
\caption{The performance of more advanced SLM on more complex dataset GSM8K \citet{gsm8k}.}
\label{tab:llama_ablation}
\end{table}

\begin{table}[t]
\centering
\renewcommand\arraystretch{1.5}
\resizebox{0.90\linewidth}{7mm}{
\begin{tabular}{cccccc}
\hline  
                                        & CoT-Finetuing & Scott & Step-by-Step & ICoT-SI   & Ours  \\ \hline
Training Time(Hour)                     &  11       & 16 & 16        & 10     &   23     \\  \hline
Training GPU Usage (G)                  &  8        & 13 & 13        & 8      &   6      \\  \hline
\end{tabular}
}  
\caption{Comparison of the average training costs required for different distillation strategies across all datasets we used. The student SLM here is GPT2-base.}
\label{tab:training_time}
\end{table}

\section{Analysis for Non-reasoning Chunks}
\label{sec:nonreasoningstas}
We demonstrate the benefits of excluding non-reasoning chunks (e.g., transitional or summary chunks) from the learning of reasoning chunks after chunking, from two perspectives. 

First, we conduct a qualitative analysis, where a case and its chunking result after SBC are shown in the Table \ref{tab:chunk result Analysis}. The core reasoning tokens in chunk 1 and chunk 2 are "17 + 10 = 27" and "27 + 35 = 62", while chunk 3 contains no core reasoning tokens, as it serves solely as a summary. Therefore, when excluding non-reasoning chunks from influencing the learning of core reasoning tokens and training reasoning chunks independently, the average share of core reasoning token in reasoning chunks increases compared to their share in the complete rationale.

Then, as shown in the Table \ref{tab:rate comparasion}, we randomly sampled 50 chunking cases from the AddSub dataset and computed the average proportion of core reasoning tokens in both the complete rationale and the reasoning chunks. The result quantitatively demonstrates the increase in the average proportion of core reasoning tokens within the reasoning chunks.

\section{Naive Method for Oversmoothing}
\label{sec:naive}
There are two naive solutions to solve the oversmoothing problem, namely weighted and refined rationale. Specifically, the first solution involves increasing the loss weight for core reasoning tokens in the rationale, while the second solution focuses on designing prompts to guide the LLM in generating refined rationales with minimal non-reasoning content.

In this work, we evaluate the feasibility of these two solutions using the Track Shuffled Objects (TSO) dataset. For the weighted solution, we leverage tokens from key exchanging results in every step as the most core reasoning tokens in the rationale. Subsequently, the loss weight for these core tokens is doubled compared to the remaining tokens. For the refined rationale solution, we design prompts (shown in the Table \ref{tab:naiveprompt}) to guide the LLM \textit{GPT-3.5-Turbo} in generating the most concise rationales. 

The results of both solutions are presented in the Table \ref{tab:naive}. The results indicate that the weighted solution performs similarly to the baseline, suggesting its effectiveness is limited. Moreover, even if this solution exhibits some effectiveness, its applicability is limited, as not all tasks can identify core reasoning tokens through artificial rules, as in TSO. The refined rationale solution demonstrates effectiveness for smaller model sizes. However, for larger model sizes, the reduced information content compared to normal rationales leads to overfitting, resulting in performance inferior to the baseline.

\section{Sentence-wise and step-wise training}
\label{sec:sentandstep}
In addition to partitioning into a fixed number of chunks, we also segment the rationale by sentences or reasoning steps, enabling the SLM to learn only one sentence or reasoning step per training iteration. For both approaches, we evaluate two schemes: one incorporating prefixes like the CWT with AC and one without prefixes. The detailed results of these approaches on the TSO dataset are presented in the Table \ref{tab:step}. As shown, the performance of all approaches exhibits a decline. The Table \ref{tab:stepbad case} also highlights the most frequent failure cases for these schemes. It can be observed that these schemes often generate repetitive reasoning steps until the maximum generation length is reached. This occurs because the number of chunks resulting from sentence- or step-based segmentation is typically variable, making it challenging for the SLM to determine the required number of reasoning steps for different problems after chunk-wise training.

\begin{table*}[hb]
\centering
\renewcommand\arraystretch{1.2}
\begin{tabular}{cccccc}
\hline
                         & Base & Base w. sent & Base w. sent prefix & Base w. step & Base w. step prefix \\ \hline
TSO                     &  37.33 & 7.11  &   27.11   & 14.22 &  30.22   \\  \hline
\end{tabular}
\caption{The accuracy of training the SLM using rationale partitioning methods with varying granularities.}
\label{tab:step}
\end{table*}

\begin{table*}[ht]
\renewcommand\arraystretch{1.2}
\begin{tabular}{p{\linewidth}}
\hline
Question: Alice, Bob, and Claire are dancers at a square dance. At the start of a song, they each have a partner: Alice is dancing with Ophelia, Bob is dancing with Rodrigo, and Claire is dancing with Patrick. Throughout the song, the dancers often trade partners. First, Bob and Alice switch partners. Then, Claire and Bob switch partners. Finally, Claire and Alice switch partners. At the end of the dance, Bob is dancing with  Which choice is true? Answer choices: (A) Rodrigo. (B) Ophelia. (C) Patrick. \\

Rationale: Sure, let's break it down step by step. At the start of a song, Alice is dancing with Ophelia, Bob is dancing with Rodrigo, and Claire is dancing with Patrick. After the first partner switch, Bob is now dancing with Ophelia and Alice is dancing with Rodrigo. After the secend partner switch, ..., After the third partner switch, ..., After the third partner switch, ..., After the third partner switch, ... \\ \hline
\end{tabular}
\caption{The base case for sentence-wise and step-wise training.}
\label{tab:stepbad case}
\end{table*}


\section{Extension of Ablation study}
\label{sec:extablation}
We further conduct extensive ablation experiments on SLMs with varying parameters. The results are presented in the Table \ref{tab:extablationbase} and Table \ref{tab:llama_ablation}. The results demonstrate that the proposed strategy performs effectively across various conditions.

\section{Extension of Comparison with Other Methods}
\label{compare_other}
The training time and GPU memory overhead of different strategies is shown in the Table \ref{tab:training_time}. 

Since chunking reduces the context length processed during each forward propagation, CWT and STT offer a unique advantage during training, that is, they require less GPU memory compared to other methods.

The increase in training time primarily results from the data chunking strategy. First, assuming a chunk contains $k$ sentences, SBC involves $M$ allocation steps, with each step generating sentence combinations at a complexity of $O(k)$, resulting in a total complexity of $O(kM)$ for SBC. In addition, STT involves the removing operater to M chunks when constructing training data, whose complexity is $O(M)$. Although both chunking and removing are linear complexity, they will bring a little additional training time compared to the baseline. Second, chunking increases the amount of training data. In theory, compared to other methods, CWT requires approximately $M+1$ times the training time, while STT requires about $num_i \times (M+2)$ times the training time, where $num_i$ refers to the iteration number for performing STT. In practice, the model does not require such an extensive amount of data to converge. Therefore, we apply an early stopping strategy, terminating training if accuracy does not improve for 10 consecutive epochs within iterations or STT achieves no improvement compared with the previous iteration. Under this setting, the proposed method requires approximately twice the training time compared to other methods.

\section{Case study}

\subsection{Core reasoning tokens}
\label{sec:corecase}
\begin{figure*}[h]
\centering
\setlength{\belowcaptionskip}{-0.5cm}
\setlength{\belowcaptionskip}{-0.5cm}
\subfigure[case 1.]{
\includegraphics[width=1\linewidth]{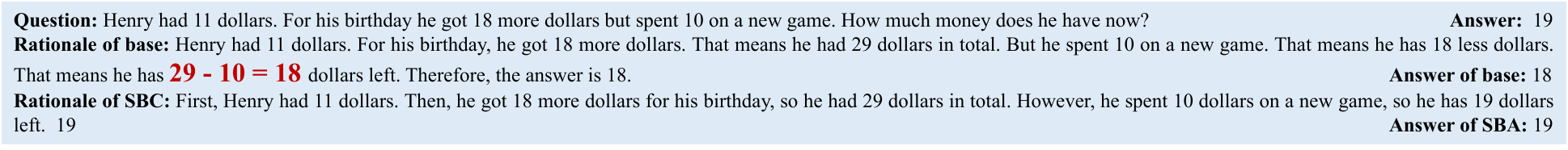} 
}
\subfigure[case 2.]{
\includegraphics[width=1\linewidth]{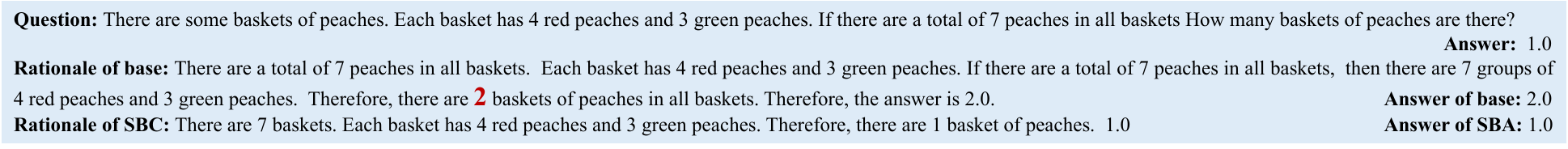} 
}
\subfigure[case 3.]{
\includegraphics[width=1\linewidth]{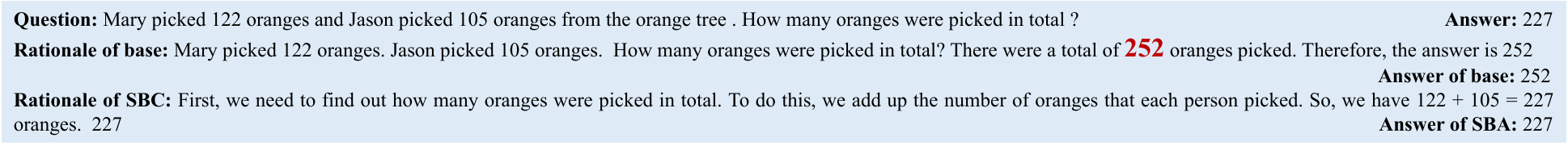}
}
\subfigure[case 4.]{
\includegraphics[width=1\linewidth]{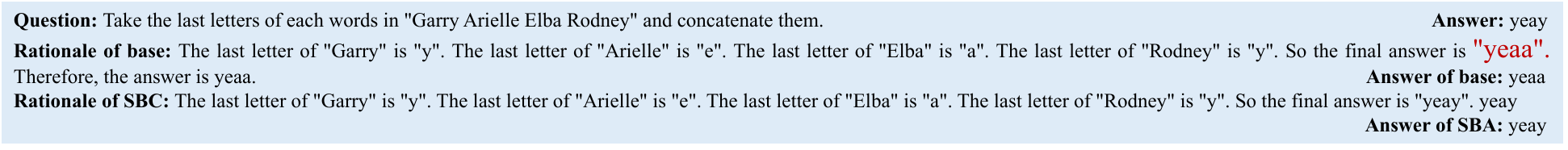} 
}
\subfigure[case 5.]{
\includegraphics[width=1\linewidth]{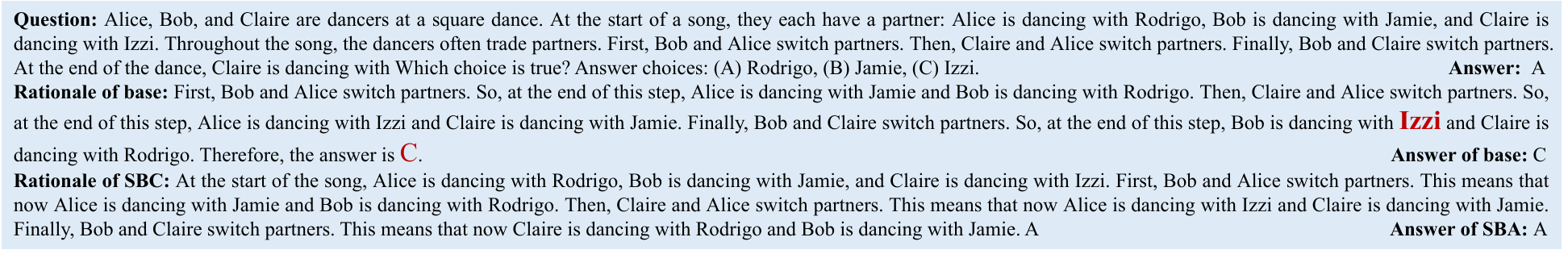} 
}
\subfigure[case 6.]{
\includegraphics[width=1\linewidth]{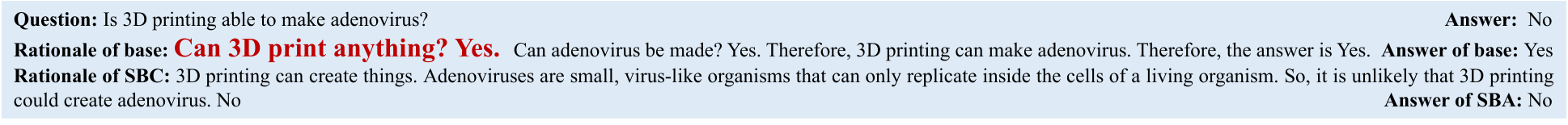}
}
\DeclareGraphicsExtensions.
\caption{The case for core reasoning tokens.}
\label{sec: core_case}
\end{figure*}

Figure \ref{sec: core_case} presents six cases across different types of benchmarks, demonstrating the improvement in the SLM's core reasoning logic following CWT training.

\subsection{Comparsion between AC and SBC}
\label{sec:SBCcase}
\begin{figure*}[h]
\centering
\subfigure[case 1.]{
\includegraphics[width=1\linewidth]{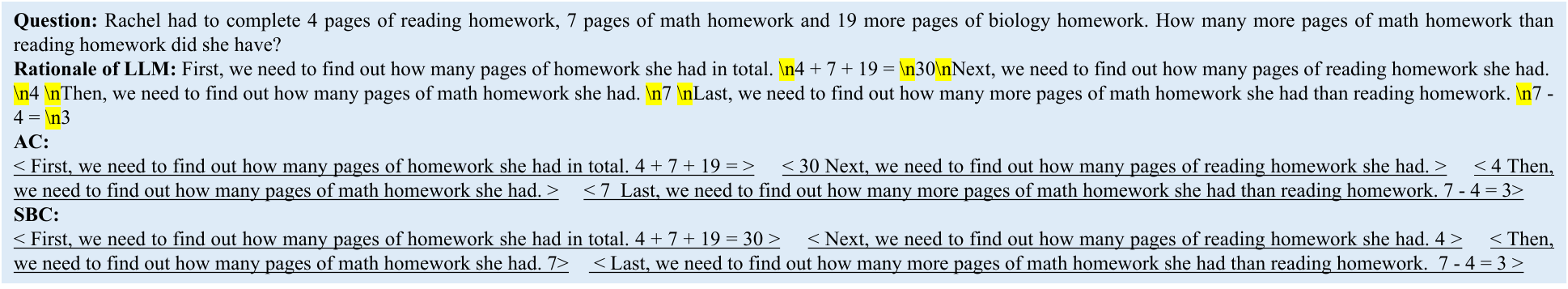} 
}
\subfigure[case 2.]{
\includegraphics[width=1\linewidth]{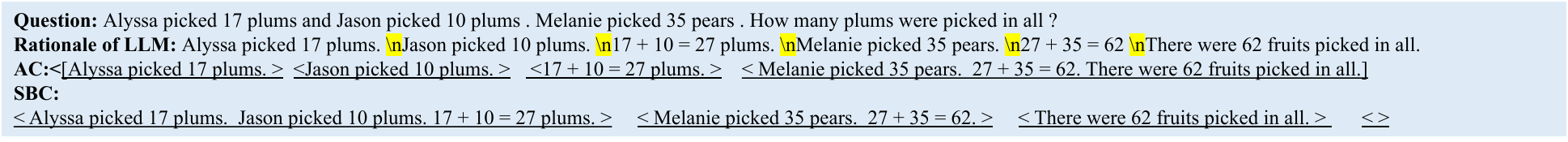} 
}
\subfigure[case 3.]{
\includegraphics[width=1\linewidth]{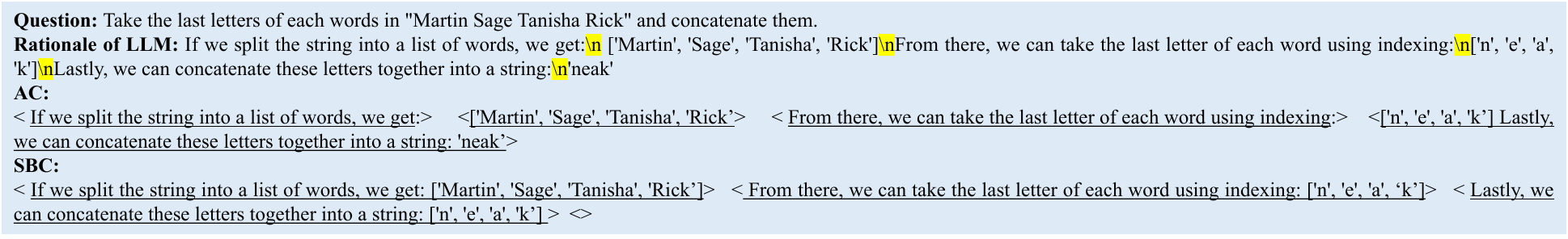}
}
\subfigure[case 4.]{
\includegraphics[width=1\linewidth]{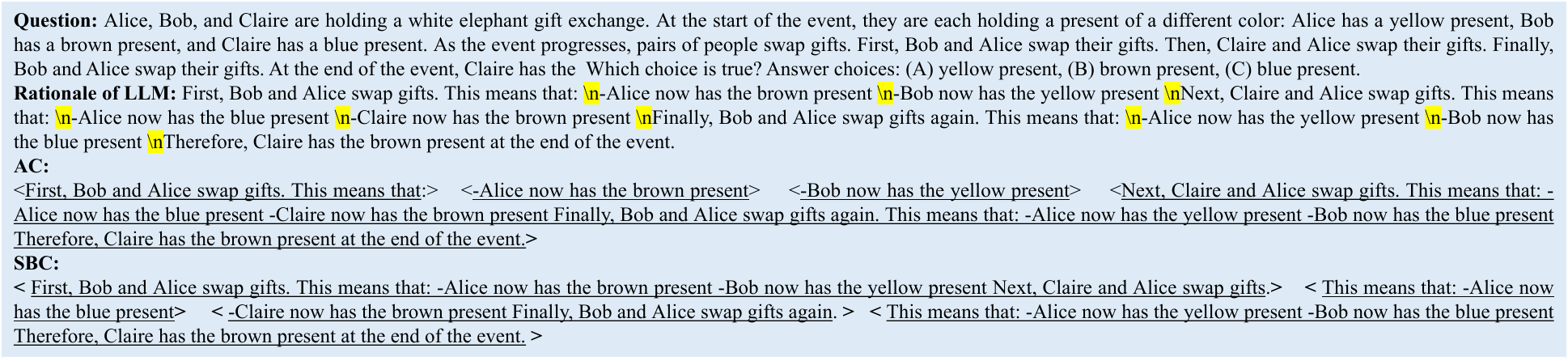} 
}
\DeclareGraphicsExtensions.
\caption{The case for SBC. <·> represents a chunk.}
\label{fig: SBC_case}
\end{figure*}

The Figure \ref{fig: SBC_case} illustrates the differences in chunk division results between AC and SBC. As shown, the SBC division results in chunks with more coherent internal semantics.

\subsection{The case for skip-thinking.}
\label{sec:core_skip}
\begin{figure*}[h]
\centering
\setlength{\belowcaptionskip}{-0.5cm}
\subfigure[case 1.]{
\includegraphics[width=1\linewidth]{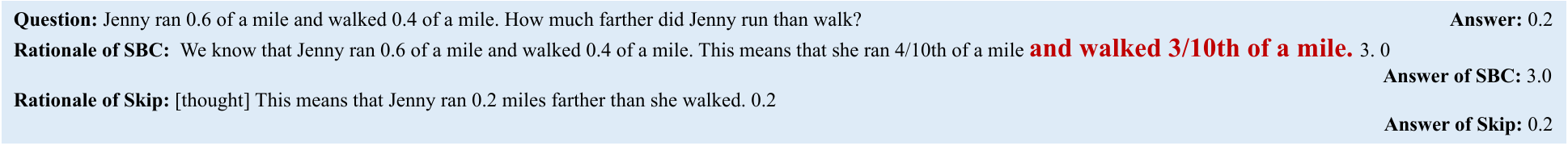} 
}
\subfigure[case 2.]{
\includegraphics[width=1\linewidth]{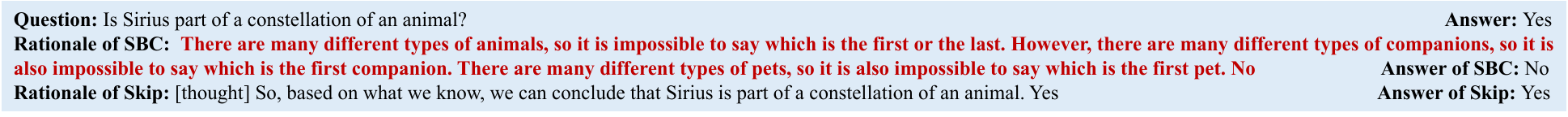} 
}
\DeclareGraphicsExtensions.
\caption{The case for Skip-thinking.}
\label{fig: skip_case}
\end{figure*}

Figure \ref{fig: skip_case} demonstrates that skip-thinking reduces the risk of SLM's hallucinations in rationale generation compared to full-thinking.

\section{Avoiding local optima}
\label{sec:localoptima}
As discussed in the section \ref{limitation}, SBC may sometimes fall into local optima, which remains a limitation of this approach. However, we emphasize that, leveraging the inherent capabilities of language models and the relative stability of sentence semantics, the overall results of SBC-based chunking are at least as effective as those obtained through average chunking. This is indirectly reflected in Table \ref{tab:1} of the original paper, where the accuracy of "\textit{Base w. SBC}" is consistently greater than or equal to that of "\textit{Base w. AC.}"

Additionally, in the section \ref{limitation}, we discuss several approaches to mitigate SBC’s local optima issue, such as simulated annealing (SA). To intuitively demonstrate its effectiveness in mitigating SBC’s local optima, we integrate simulated annealing into the SBC method by introducing a temperature parameter $T=0.1$, allowing a certain probability of accepting suboptimal partitions identified by SBC to prevent getting stuck in local optima. As shown in the Table \ref{tab:SA}, simulated annealing effectively mitigates SBC’s local optima issue, leading to improved model performance. 

\begin{table*}[hb]
\centering
\renewcommand\arraystretch{1.2}
\begin{tabular}{cccccccc}
\hline
                         &SE	&AD	&MA	&Svamp&	TSO&	LLC	&SQA \\ \hline
Base w. SBC              &8.55	&10.92&	17.77&	11.33&	100.00&	25.33&	59.38         \\  \hline
Base w. SBC \&SA  &10.52	&11.76	&18.88	&12.33	&100.00	&26.00&	59.53 \\ \hline
\end{tabular}
\caption{Effect of integrating simulated annealing into SBC. The student SLM is GPT2-base.}
\label{tab:SA}
\end{table*}

\end{document}